\documentclass[letterpaper, 10 pt, conference]{ieeeconf}  % Comment this line out if you need a4paper

\IEEEoverridecommandlockouts                              % This command is only needed if 
\usepackage{graphicx}
\usepackage[sort,compress]{cite}
\usepackage{xcolor,balance}
\usepackage{todonotes}
\usepackage{amsmath}
\usepackage{amssymb}
\usepackage{subcaption}
\usepackage[font=footnotesize]{caption}

\usepackage{multirow}
\usepackage{multicol}

\def\argmin{\mathop{\rm argmin}}

\title{\LARGE \bf Multi-Agent Relative Pose Estimation with UWB\\% 
and Constrained Communications}
\author{Andrew Fishberg \and Jonathan P. How
\thanks{* Work supported in part by DOE, NNSA, and ALB funding. Both authors are with the MIT Department of Aeronautics and Astronautics. Contact at \tt\{fishberg, jhow\}@mit.edu}%
}

\date{}

\begin{document}
% Force the use of "et al." for references with too many authors
\bstctlcite{bst_control}

\maketitle

\begin{abstract}
Inter-agent relative localization is critical for any multi-robot system operating in the absence of external positioning infrastructure or prior environmental knowledge. We propose a novel inter-agent relative 2D pose estimation system where each participating agent is equipped with several ultra-wideband (UWB) ranging tags. Prior work typically supplements noisy UWB range measurements with additional continuously transmitted data, such as odometry, making these approaches scale poorly with increased swarm size or decreased communication throughput. This approach addresses these concerns by using only locally collected UWB measurements with no additionally transmitted data. By modeling observed ranging biases and systematic antenna obstructions in our proposed optimization solution, our experimental results demonstrate an improved mean position error (while remaining competitive in other metrics) over a similar state-of-the-art approach that additionally relies on continuously transmitted odometry.
\end{abstract}

%As a planned component of a larger communication constrained multi-agent SLAM pipeline, 
\section{Introduction}
\label{sec:intro}
Multi-robot approaches improve the efficiency and robustness of large-scale decentralized tasks such as search \& rescue \cite{DBLP:journals/ijrr/TianLOTARH20}, warehouse automation \cite{DBLP:journals/computers/LiuCYJ18}, and planetary exploration \cite{DBLP:journals/corr/abs-2103-11470}. To operate and divide tasks effectively, all these systems require an understanding of where agents and their peers are within a common reference frame. In practice this is often achieved by localizing within an a priori map or using an external measurement system like GPS or motion capture (mocap). In scenarios where these convenient technologies are unavailable or infeasible, common approaches utilize both relative localization \cite{DBLP:journals/trob/ZhouR08a} and decentralized SLAM \cite{DBLP:conf/icra/ChangTHC21,DBLP:journals/ral/LajoieRCCB20,DBLP:conf/icra/CieslewskiCS18} techniques.

Relative localization is often computed from a set of relative range or relative angle measurements. In turn, these inputs are estimated by measuring a received signal's time of arrival (TOA), time difference of arrival (TDOA), angle of arrival (AOA), or received signal strength (RSS) \cite{li2016contributed}. Furthermore, said signals have many possible forms: acoustic \cite{DBLP:journals/ral/FischellRS19}, Bluetooth Low Energy (BLE) \cite{DBLP:journals/iotj/JeonSSN18}, Radio Frequency Identification (RFID) \cite{sanpechuda2008review}, or WiFi \cite{DBLP:conf/icra/HerranzLMO14,arun2022p2slam}. Within the last decade, ultra wideband (UWB) has matured into a reliable, inexpensive, and commercially available RF solution for data transmission, TOA or TDOA ranging, and localization. 
As a result, many roboticists have already begun to incorporate UWB into their work (see Section \ref{sec:relwork}).
Several advertised properties make UWB particularly noteworthy: precision of approximately $10$ centimeters, ranges up to $100$ meters, resilience to multipath, no dependency on line of sight (LOS), low power consumption, and $100$ Mbit/s communication speeds \cite{DBLP:journals/sensors/AlarifiAAAAAA16}. Nevertheless, UWB measurements are not immune from ranging errors or noise, the modeling and correction of which is an active research topic within the robotics community \cite{DBLP:conf/comsnets/SmaouiGK20,DBLP:conf/ccta/LedergerberD17,DBLP:journals/access/LedergerberD18,DBLP:journals/access/HamerD18}.
A common approach among UWB relative localization work supplements noisy UWB ranging measurements with additional continuously transmitted data, such as odometry \cite{DBLP:conf/iros/Cao0YANMT21} and visual inter-agent tracks \cite{DBLP:conf/icra/XuWZQS20,DBLP:journals/corr/abs-2103-04131}. While this additional data improves overall estimation accuracy, its transmission causes poor scalability with respect to increased swarm size or decreased communication throughput.

\begin{figure}[t]
    \centering
    \includegraphics[width=\linewidth*8/10]{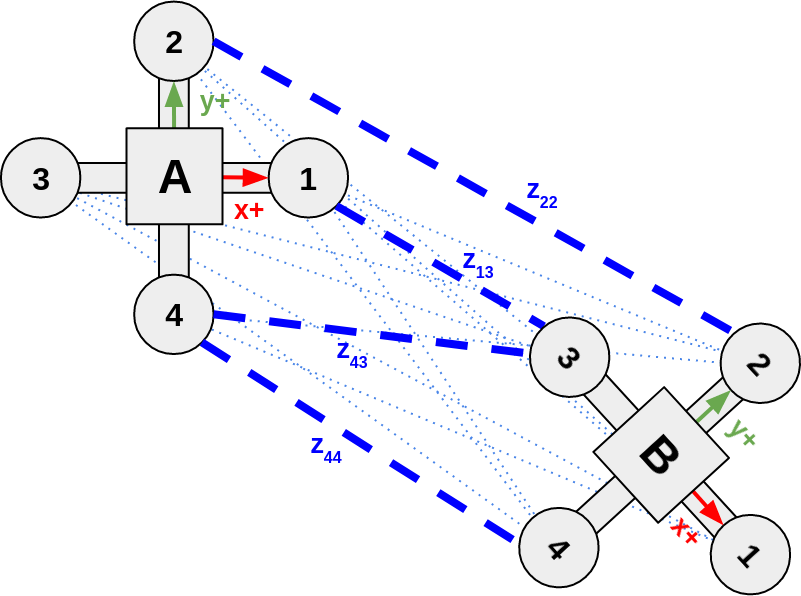}
    \caption{Diagram of two agents, \texttt{RobotA} and \texttt{RobotB}, each with their respective relative coordinate frames and UWB sensors labeled. Relative range measurements are shown between each pair of inter-agent antennas in blue (N.B.: the dashed and dotted blue lines are only differentiated for the sake of visual clarity, they represent the same type of measurements). These 16 unique measurements are denoted as $z_{ij}$ where $i$ and $j$ are the corresponding antenna indices for \texttt{RobotA} and \texttt{RobotB} respectively.}
    \label{fig:coordinates}
\end{figure}

% multi-agent

This paper presents a multi-tag UWB relative 2D pose estimation system built for future integration into a larger resource-aware distributed SLAM pipeline. Specifically, our system aims to produce collaborative inter-agent loop-closures (i.e. relative pose estimates between agents) at a higher rate than visual loop-closure techniques utilized by decentralized visual SLAM systems \cite{DBLP:conf/icra/ChangTHC21}, but with a lower communication footprint and absolute accuracy. We achieve this by equipping each agent with multiple UWB tags in a known prior configuration (such as shown in Figure \ref{fig:coordinates}), allowing our system to produce full relative pose estimates between agents using only locally collected UWB measurements (i.e. no inter-agent data, such as odometry, is exchanged). By using only local measurements, the host distributed SLAM system can allocate its full limited bandwidth to the transmission and detection of the higher fidelity visual loop-closures.\footnote{As the resource-aware detection of collaborative loop-closures is a research topic in itself \cite{DBLP:journals/ijrr/TianKH21}, how the host decentralized SLAM pipeline can best utilize these UWB loop-closures remains a subject for future work. Promising potential directions include: (1) to improve the state estimation in the absence of visual loop-closures and (2) to inform the priority of transmitting potential loop-closure between agents.}

Our approach is distinguished from similar UWB-enabled relative localization works by completely forgoing the use of transmitted (remote) measurements, 
%in turn addressing the scalability concerns.Our approach addresses these concerns by consciously restricting itself to locally collected UWB measurements, 
effectively trading off a small reduction in absolute estimation accuracy for a superior (i.e., eliminated) communication footprint and thus scalability.
Despite this potential performance sacrifice, by accounting for the measurement errors (i.e., systemic UWB ranging biases and systematic obstructions) in the optimization process, we achieve superior mean position accuracy and comparable performance on other metrics to prior work \cite{DBLP:conf/iros/Cao0YANMT21} without the need to continuously transmit odometry estimates.

This paper's contributions are:
(1)~An in-depth analysis and modeling of the observed noise characteristics of UWB ranging measurements (Section \ref{sec:characterizing-noise});
(2)~A customized solution for UWB-based relative localization that takes into account the specific observed noise properties, formulated as a nonlinear least squares (NLLS) optimization (Section \ref{sec:problem}); and
(3)~Experimental results that demonstrate the merits of the developed solution (Section \ref{sec:results}).

\section{Related Work}
\label{sec:relwork}

\begin{figure}[t]
    \centering
    \begin{subfigure}{.50\linewidth}
    \includegraphics[width=\linewidth]{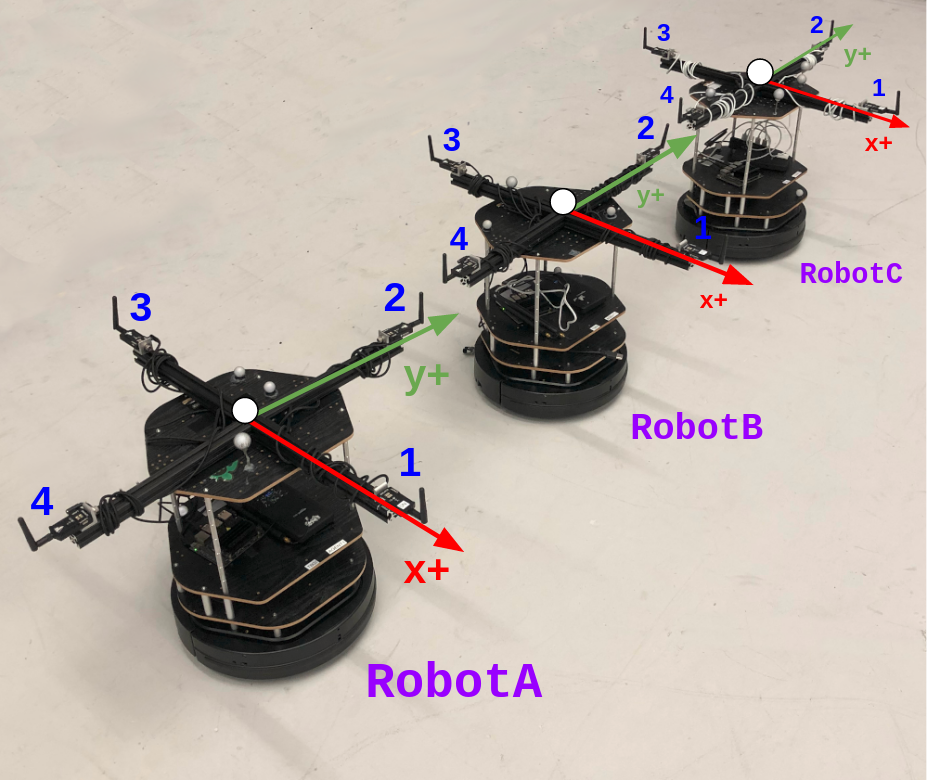}
    \caption{}
    \label{fig:robot}
    \end{subfigure}
    \hfill
    \begin{subfigure}{.42\linewidth}
    \includegraphics[width=\linewidth]{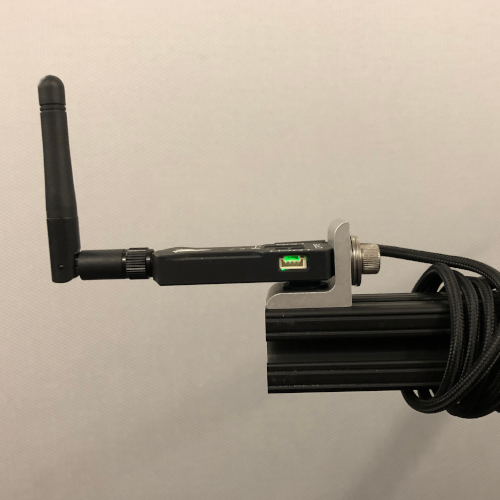}
    \caption{}
    \label{fig:mount}
    \end{subfigure}
    \hfill
    \caption{(a) Photo of our experimental robots. Each is comprised of a Turtlebot2 base, a Jetson Nano, and four Nooploop UWB sensors. Robots are manually controlled with a wireless Logitech Gamepad. (b) A close up showing how UWB sensors are attached radially from the aluminum mounting bracket. This is done to ensure antenna clearance and to avoid direct contact with metal.}
    \label{fig:hardware}
\end{figure}

%Previous work has examined a wide range of SLAM and localization techniques leveraging various types of relative ranging measurements such as BLE, WIFI, and sonar. 
%Robotic localization with range measurements spans multiple mediums (e.g. GPS, UWB, BLE, RFID, WiFi, sonar, etc.) and techniques.
The recent paper \cite{DBLP:conf/meco/YuLQHW21} provides an excellent overview recent UWB usage in robotics and IoT pipelines. When considering UWB for mobile robotics, the work naturally
separates into several categories.

\noindent
\textbf{Known anchor points:}
In these systems a target with a single UWB tag is trilaterated by several UWB anchor points at known global locations. These approaches are conceptually similar to GPS where each range measurement is coming from a satellite with a known absolute position, although anchors are often assumed to be static. Since the target has a single tag, the target's orientation is not observable from just UWB measurements.
Works using these assumptions produce improved state estimation results by fusing the global UWB position estimation with some combination of dynamics models, odometry (e.g. wheel odometry or visual odometry), and IMU data \cite{DBLP:journals/access/StrohmeierWRM18,yang2021uvip,DBLP:conf/icra/FontanelliSBP21,DBLP:journals/ral/NguyenNX21,DBLP:conf/icra/MullerHD15}. 

%Many of the most popular commercially available UWB systems (i.e. MDEK1001 Development Kit, Pozyx Developer Tag, and Nooploop LinkTrack P) advertise this as their default operation mode, and consider it a cheap alternative to mocap system.

\noindent
\textbf{Static features:}
These works assume there are one or more static UWB anchors with not necessarily known positions. The observed anchor(s) are treated like static SLAM feature(s) and fused with odometry data to perform localization \cite{DBLP:conf/icra/HerranzLMO14,DBLP:journals/corr/abs-1804-09773,DBLP:journals/corr/abs-2109-00747}. This is also comparable to \cite{arun2022p2slam}, but WiFi hotspot relative bearing measurements are used instead of UWB ranging measurements. Also in this category is \cite{DBLP:journals/ral/FunabikiMNA21,DBLP:journals/corr/abs-2103-11470}, where the robot deploys a stationary UWB node during exploration, effectively marking a static point in the environment useful for future loop-closure detection.

\noindent
\textbf{Observability results:}
Mobile agents, each equipped with one (or possibly two in the case of \cite{DBLP:journals/ral/ShalabyCFN21}) ranging tags, take several temporally spaced relative range measurements while each traversing some trajectory. Agents track their local trajectory through odometry or IMU dead reckoning and share it with other agents. As shown in \cite{DBLP:journals/trob/ZhouR08a,DBLP:journals/ral/CossetteSSFN21}, if the trajectories have sufficient relative motion then the relative pose between agents can be recovered. 
%This is also includes , which establishes similar observability conditions for a swarm of quadcopterseach equipped with one or two UWB tags.
%are non-degenerative (i.e
%3D system with agents having either one or two tags. Uses IMU and Multiplicative Extended Kalman Filter (MEKF) to estimate positioning (not heading). Has observability analysis for required trajectory and recommends grounding certain vehicles with GPS (end of Section III). Using Pozyx. Has good paper references in Introduction. \cite{DBLP:journals/ral/ShalabyCFN21}

\noindent
\textbf{Multi-tag pose estimation:}
Mobile agents are each equipped with multiple non-collocated tags at known relative positions on the agent. With sufficient tags, relative position and orientation can be observed with UWB measurements alone \cite{DBLP:conf/intenv/BonsignoriCGGLM20,DBLP:conf/vtc/GhanemOK20,DBLP:conf/i2mtc/TheusslNO19}. Furthermore, additional measurements, such as odometry, optical flow, IMU readings, and altitude measurements can be communicated between agents and fused with the UWB measurements in \cite{DBLP:conf/iros/Cao0YANMT21} and \cite{DBLP:conf/icra/NguyenZWGX18} to improve results and achieve observability respectively.
Ref.~\cite{DBLP:conf/iros/Cao0YANMT21} presents similar work to this paper, but it differs in the particle filtering approach and by the need to continuously share odometry between agents which contrasts with the objectives of this work. The similarities between the hardware setup and experiments allow for a useful baseline of comparison for our experimental results (see Section \ref{sec:comparison}).

\noindent
\textbf{Correcting UWB ranging errors:} 
Works in this category model the ranging errors or design calibration schemes to correct errors
\cite{DBLP:conf/comsnets/SmaouiGK20,DBLP:conf/ccta/LedergerberD17,DBLP:journals/access/LedergerberD18,DBLP:journals/access/HamerD18}. Additionally, as many of the commercially available UWB sensors provide ranging measurements as a black box, certain algorithmic choices and calibration parameters are hidden from the user. %Sometimes, as is the case in this work, a layer of signal/error modeling is laid over the closed source product to address observed arbitrary sensor quirks.

\noindent
\textbf{Full pipelines:}
Omni-swarm \cite{DBLP:journals/corr/abs-2103-04131} (and earlier work \cite{DBLP:journals/trob/QinLS18,DBLP:conf/icra/XuWZQS20}) is perhaps the most complete and comprehensive UWB localization system. However, the impressive results achieved by the system heavily rely on omni-directional cameras for tracking neighboring agents and, although decentralized, mandates a continuous exchange of odometry and other measurements between all agents. Additionally, \cite{steup2021single} provides an interesting full pipeline approach, but has each agent's role alternate between active swarm member and static anchor node.

%Thus, this pipeline is out of our application scope.
%Given our communications and scalability focus, this
\section{Characterizing UWB Noise}
\label{sec:characterizing-noise}

\subsection{Robot and UWB Setup}
\label{sec:hardware-robot}
All experiments were conducted with a set of robots (see Figure \ref{fig:robot}) each equipped with a Turtlebot2 base, NVIDIA Jetson Nano, and four Nooploop UWB sensors. UWB sensors are positioned $R = 0.35$ m from the center of the robot to the upright UWB antenna.\footnote{As the robustness and accuracy of the estimated pose computed from trilateration is positively correlated with the length of the sensor baseline the $R$ separation was selected to closely resemble the arm length of a medium-sized quadrotor, in anticipation for future work, as well as allowing for a fair comparison to the results of \cite{DBLP:conf/iros/Cao0YANMT21}, who tested a similar baseline.} The mounting bracket is made of aluminum and UWB modules are mounted radially for antenna clearance (see Figure \ref{fig:mount}).

In accordance to the LinkTrack manual \cite{linktrack-manual}, all antennas are all positioned upright, since their $xy$-plane has better omni-directivity than their $z$-axis. UWB nodes are numbered as in Figure \ref{fig:coordinates}. Nodes are configured into \textit{Distributed Ranging (DR) Mode}, to enable measuring relative range without the need for stationary anchors, and configured to use Channel 3 (3,744-4,243.2 MHz) \cite{linktrack-datasheet}. All tests were performed in a motion capture space for high-precision ground truth pose.

The Nooploop LinkTrack P was selected over other considered products, such as the Pozyx Developer Tag or DWM1001 Development Board, because: (1) the LinkTrack P has a slim form factor (5.5cm x 3cm x 0.75cm), light weight (33g), and easy mounting via a 1/4-20 screw hole; (2) Nooploop provides an out of the box ROS driver; and (3) in our trials using 8 tags, the LinkTrack P provided measurements at steady 50Hz, while the Pozyx Developer Tag fluctuated around 5Hz.

%Multiple commercially available off-the-shelf relative-ranging UWB solutions are now available. Several options were purchased and tested in a laboratory setting. Fully functioning localization systems were produced with the Pozyx (based on Decawave's DW1000 IC) and Nooploop products. Both provided similar ranging accuracy and measurement noise, but the 

%Most of the US and Europe products are based on the Decawave integrated circuits (i.e. DW1000 IC, DW3000 IC).
%(3) Multiple relevant published works in this space %use the LinkTrack P to great effect \ref{papers} %\mytodo{add ref}, allowing for a fairer comparison of %results.

\subsection{Range Measurement Bias \& Noise}
\label{sec:measurement-bias-noise}

\begin{figure}[t]        \captionsetup[subfigure]{aboveskip=-1pt,belowskip=-1pt}
    \centering
    \hfill
    \begin{subfigure}{\linewidth*3/7}
    \includegraphics[width=\linewidth]{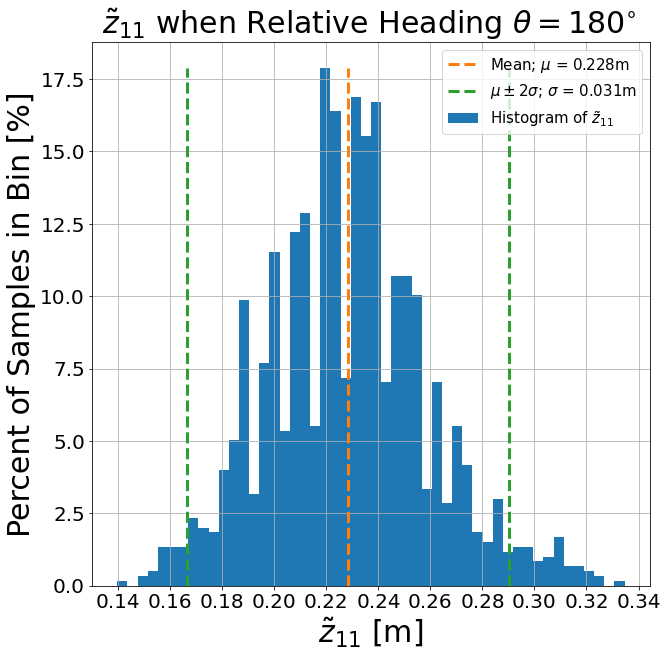}
    \caption{}
    \label{fig:histogram-180}
    \end{subfigure}
    \hfill
    \begin{subfigure}{\linewidth*3/7}
    \includegraphics[width=\linewidth]{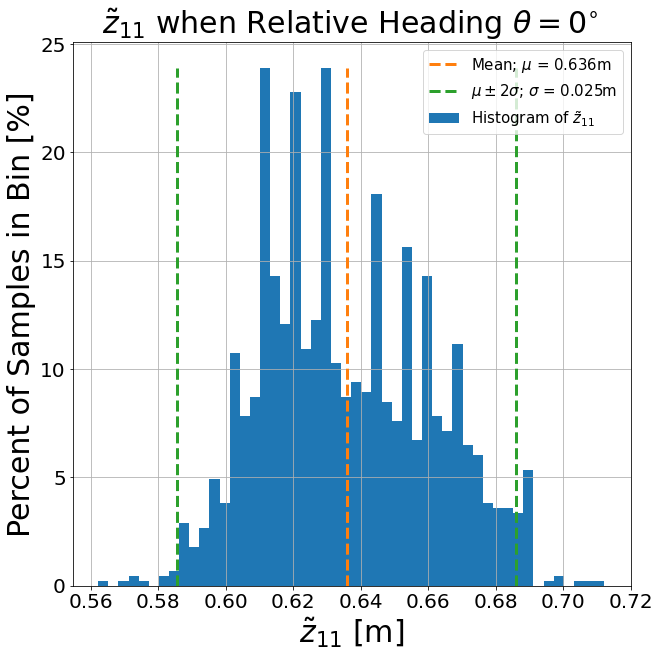}
    \caption{}
    \label{fig:histogram-000}
    \end{subfigure}
    \vspace{.2cm}
    \hfill
    \begin{subfigure}{\linewidth*3/7}
    \includegraphics[width=\linewidth]{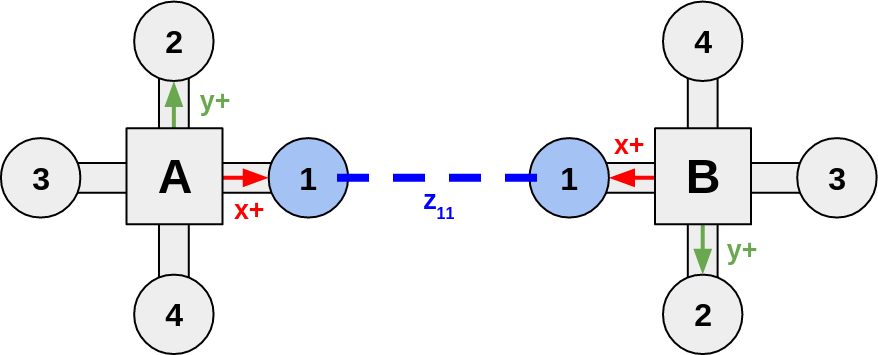}
    \caption{}
    \label{fig:z11-los}
    \end{subfigure}
    \hfill
    \begin{subfigure}{\linewidth*3/7}
    \includegraphics[width=\linewidth]{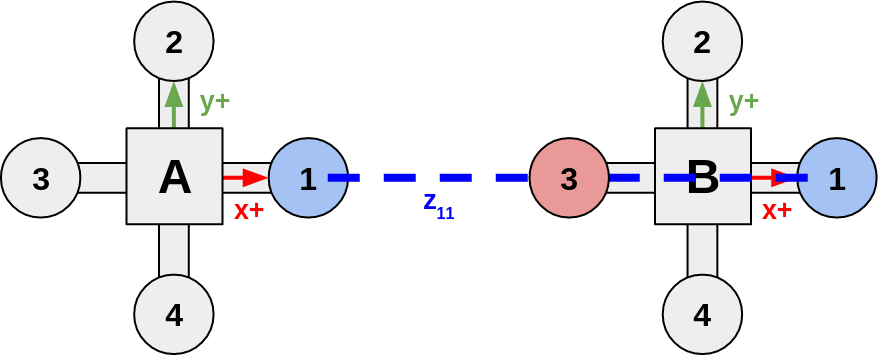}
    \caption{}
    \label{fig:z11-obstructed}
    \end{subfigure}
    \caption{Histograms showing the measurement error for two separate data collects of $\tilde{z}_{11}$ ($50$Hz for $30$ s). Measurement error $\tilde{z}_{11}$ is the difference between the UWB range measurement and ground truth distance as measured by a Vicon system.
    Histogram (a)/(b) correspond to the error from the direct line-of-sight/obstructed configuration shown in (c)/(d) respectively.
    Neither of these distributions would be well-modeled as Gaussian.
    The orange lines and green lines show the calculated sample mean and $\pm$ two standard deviation boundaries respectively. With no obstruction, we see an average of $23$ cm over-estimation of $z_{11}$, and when obstructed, this increases by $41$ cm, to an average of $64$ cm over-estimation. Thus the error is not zero mean (and the mean increases further  when obstructed).}
    \label{fig:gaussian-noise}
\end{figure}

Experiments were performed to characterize the UWB ranging noise.
Figure \ref{fig:histogram-180} shows a sample of measurement error between two stationary UWB nodes with direct line-of-sight (LOS), as shown in Figure \ref{fig:z11-los}. Three important observations are: (1) Contrary to the common measurement noise assumption, the error appears neither zero mean nor Gaussian (i.e. shown sample fails the \texttt{scipy.stats.normaltest} function \cite{2020SciPy-NMeth}, an implementation of the D'Agostino and Pearson's normal test \cite{d1971omnibus}, with a $p$-value of $2.36\times 10^{-6}$). (2) Sensors tend to consistently over-estimate the relative distance between nodes. %(i.e. $\tilde{z}_{ij}^{(t)} > 0$).
(3) Within our operating environment, the mean error between a pair of unobstructed UWB nodes remains approximately constant independent of distance between nodes or when data was collected. Additional evidence of this claim can be seen in unobstructed portions of Figures \ref{fig:z11-err-vs-heading}-\ref{fig:z14-err-vs-heading}, a follow-on experiment motivated in Section \ref{sec:antenna}.

% reducing signal susceptibility to individual noisy measurements

\subsection{Antenna Obstruction \& Interference}
\label{sec:antenna}

\begin{figure*}         \captionsetup[subfigure]{aboveskip=-0.5pt,belowskip=-1pt}

    \centering
    \hfill
    \begin{subfigure}{0.17\textwidth}
    \includegraphics[width=\linewidth]{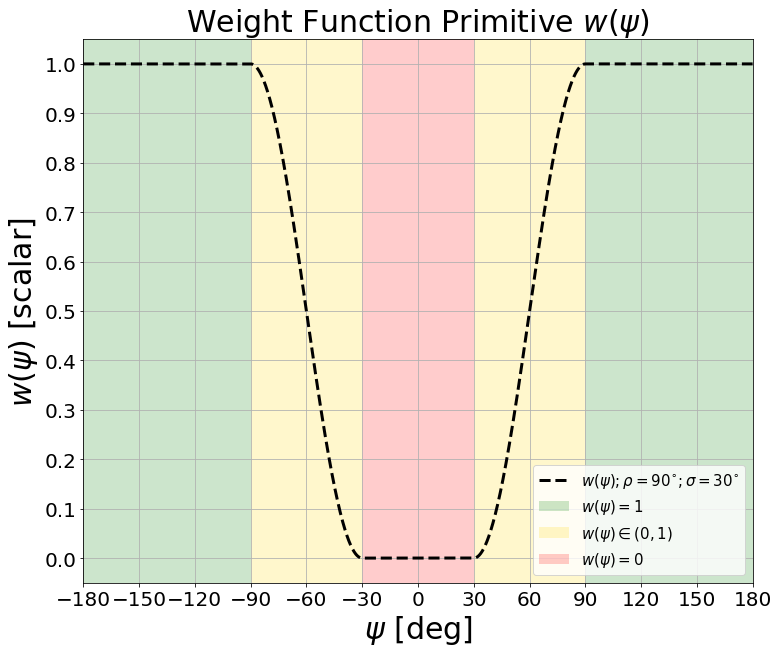}
    \caption{}
    \label{fig:weight-func}
    \end{subfigure}
    \hfill
    \begin{subfigure}{0.18\textwidth}
    \includegraphics[width=\linewidth]{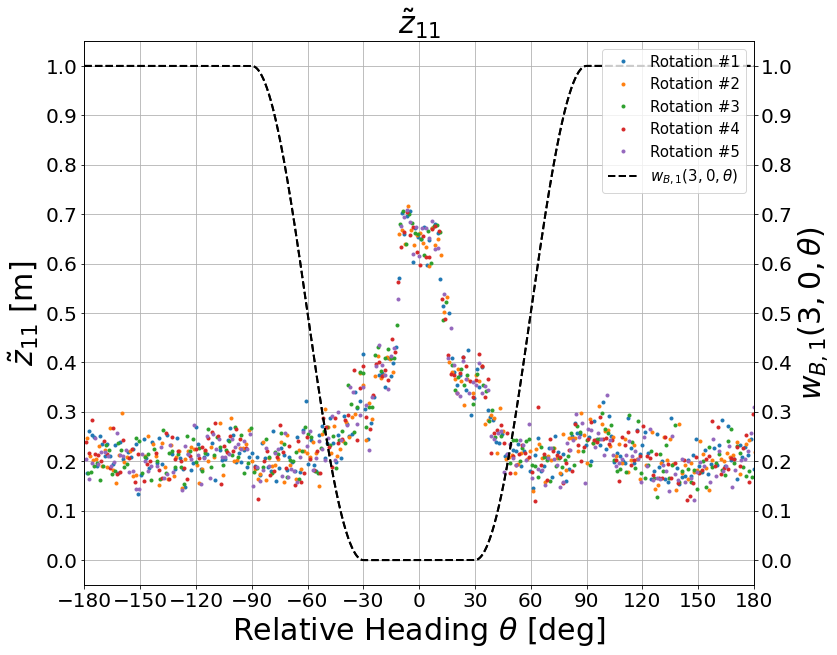}
    \caption{}
    \label{fig:z11-err-vs-heading}
    \end{subfigure}
    \hfill
    \begin{subfigure}{0.18\textwidth}
    \includegraphics[width=\linewidth]{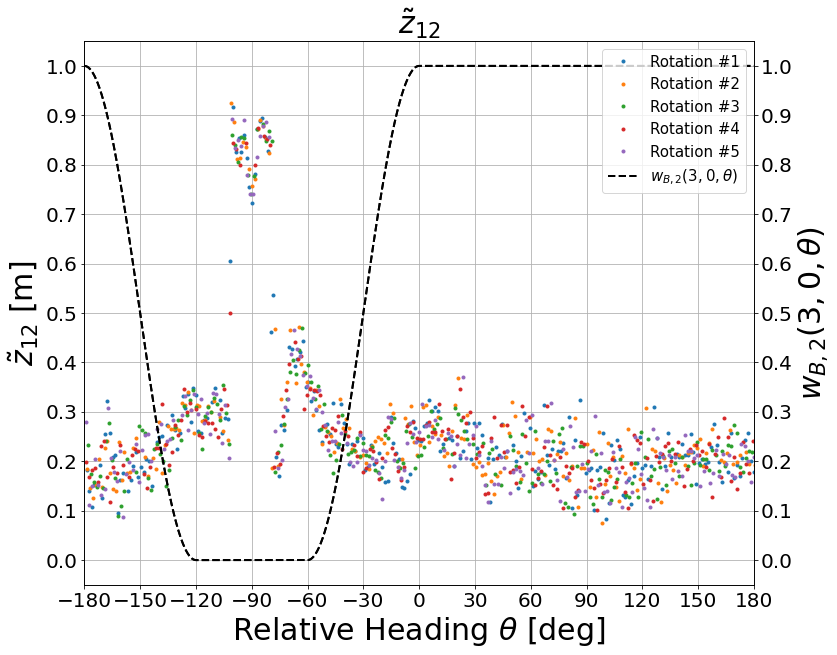}
    \caption{}
    \label{fig:z12-err-vs-heading}
    \end{subfigure}
    \hfill
    \begin{subfigure}{0.18\textwidth}
    \includegraphics[width=\linewidth]{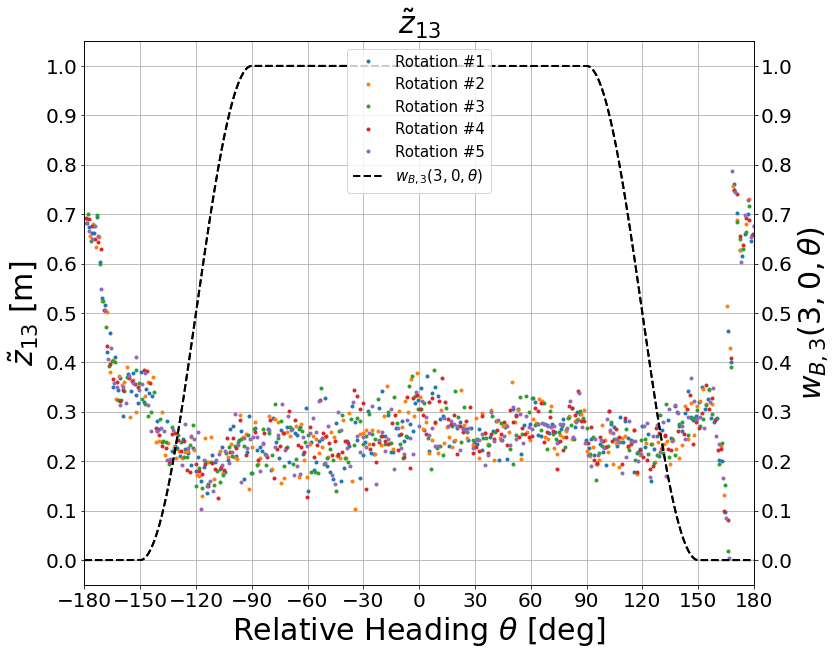}
    \caption{}
    \label{fig:z13-err-vs-heading}
    \end{subfigure}
    \hfill
    \begin{subfigure}{0.18\textwidth}
    \includegraphics[width=\linewidth]{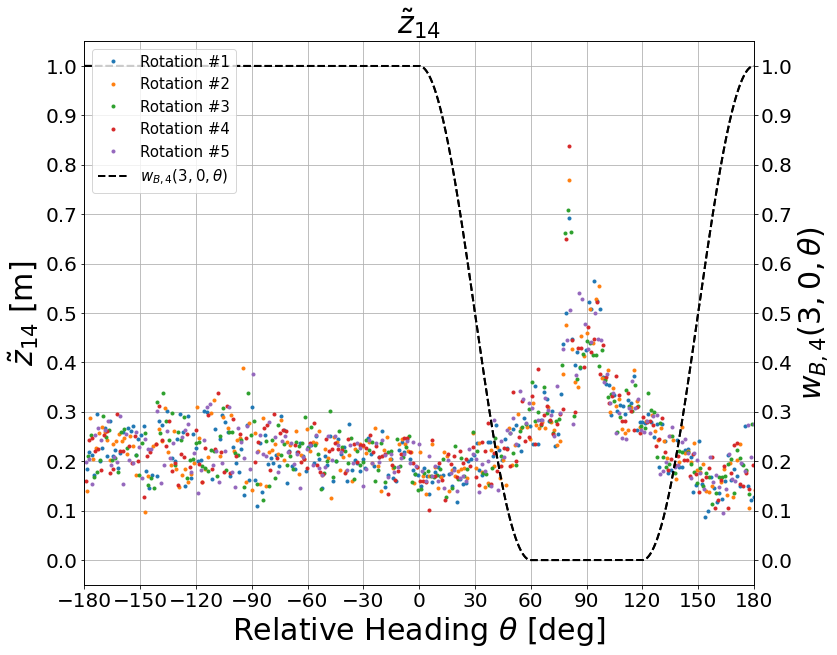}
    \caption{}
    \label{fig:z14-err-vs-heading}
    \end{subfigure}
    \hfill \vspace*{-0.05in}
    \caption{(a) Weight function primitive $w(\psi)$ with parameters $\sigma = 30^\circ$ and $\rho = 90^\circ$ (see Eq.~\ref{eq:w}). Colors represent the segments of the piecewise function, and align with the colors in the Figure~\ref{fig:antenna-ignore}.
    (b-d) Plots of $\tilde{z}_{11}$ through $\tilde{z}_{14}$ with respect to $R_B$'s relative heading. Shown data is all from the same data collection where $R_A$ was kept stationary while $R_B$ performed $5$ in place counter-clockwise revolutions at approximately $60$ deg/s. Each revolution is denoted in a different color to highlight this noise pattern is repeatable. The overlaid weight function $w_{B,j}(\mathbf{x})$ (see Eq.~\ref{eq:w_b}) shows that areas with error spikes correspond to relative headings correctly predicted and devalued by $w_{B,j}(\mathbf{x})$. Additionally, when the error is not spiking, we see additional evidence that the noise is not zero mean, as discussed in Section 
    \ref{sec:measurement-bias-noise}.}
    \label{fig:yaw-error}
\end{figure*}

Since UWB relative range measurements are a TOA measuring scheme, any delays on receiving a signal (i.e. propagating through an obstruction) will result in a ranging over-estimation. This helps make sense of the substantial $41$ cm increase of mean error shown in Figures \ref{fig:histogram-000} and \ref{fig:z11-obstructed} when compared to Figures \ref{fig:histogram-180} and \ref{fig:z11-los}. Furthermore, the LinkTrack P manual \cite{linktrack-manual} offers several notes regarding obstruction: (1) Signal can propagate through 2 or 3 solid walls, but each wall introduces approximately $30$ cm of error and a decreased maximum ranging distance. (2) The distance between each node and the obstruction affects ranging accuracy; the best results occur when the obstruction is equally spaced between the antennas, the worst results occur when the obstruction is close to one of the antennas.
%(3) Metal can absorb UWB electromagnetic waves, so close proximity can introduce measurement errors or make ranging impossible.
%
This presents a noteworthy concern: as agents move relative to each other, fixed obstructions on the robot, such as a third antenna, can pass between the ranging pair causing an obstructed measurement. This obstruction is worsened by its guaranteed proximity to at least one of the ranging antennas. This eclipse-like effect can be observed in Figures \ref{fig:z11-err-vs-heading}-\ref{fig:z14-err-vs-heading}, where clear over-estimation error spikes occur predictably at specific relative heading for each pair of antennas.

While it is difficult to pinpoint the exact cause of this obstruction, it appears to be a combination of proximity to other antennas, metal, and other components. Though it would be possible to mount these sensors in a different configuration to mitigate this, in reality there are times when obstructions cannot be avoided, especially as one begins to consider quadrotors and 3D environments. Thus, we will treat this interference as part of the given hardware setup which in turn must be mitigated algorithmically.\footnote{Inspection of the received RSSI does not appear to provide any meaningful way to detect the current obstruction. As per \cite{linktrack-manual}, when RSSI is less than $6$ dB, it is likely to be in the line-of-sight state, and when greater than $10$ dB, it is likely to be in the non-line-of-sight or multipath state. Despite this, we consistently observe \texttt{fp\_rssi} and \texttt{rx\_rssi} of approximately $-81.0$ dB and $-78.0$ db respectively, with a $0.5$ dB minimum resolution.}
\section{Optimization Formulation}
\label{sec:problem}

% \underset{\mathbf{x} = (x,y,\theta)}{\arg\min}\sum_{i \in \{1,2,3,4\}} \sum_{j \in \{1,2,3,4\}} \Big( |\texttt{ant}_i(0,0,0) - \texttt{ant}_j(x,y,\theta)|_2 - (z_{ij} - \mu_{ij}) \Big)^2 \cdot w_{base,~i}(x,y) \cdot w_{other,~j}(x,y,\theta)

Consider two robots, \texttt{RobotA} and \texttt{RobotB}, denoted as $R_A$ and $R_B$ respectively. Labels are assigned such that $R_A$ is trying to estimate $R_B$'s relative pose with respect to $R_A$'s reference frame. Let $R_B$'s relative pose be described by the state vector $\mathbf{x}$. Each robot has $N$ UWB range sensors, uniquely identified with an integer index from the set $S := \{1,\dots,N\}$. Assuming $R_A$ and $R_B$ have the same antenna layout, the relative position vector of the $k$th antenna of a robot with a pose vector of $\mathbf{T}$ is defined as $\boldsymbol{\alpha}_k(\mathbf{T})$. Since we are in $R_A$'s reference frame, $\boldsymbol{\alpha}_i(\mathbf{0})$ gives the position vector of $R_A$'s $i$th antenna, while $\boldsymbol{\alpha}_j(\mathbf{x})$ gives the position vector of $R_B$'s $j$th antenna. Thus, let $d_{ij}$ be the distance between $R_A$'s $i$th antenna and $R_B$'s $j$th antenna when $R_B$ is at $\mathbf{x}$, such that:
\begin{align}
d_{ij}(\mathbf{x}) &= \|\boldsymbol{\alpha}_i(\mathbf{0}) - \boldsymbol{\alpha}_j(\mathbf{x})\|_2
\end{align}
For each discrete timestep $t$, $N^2$ raw relative range measurements are taken, one per unique pair of $R_A$ and $R_B$ antennas. These measurements are denoted as $z_{ij}^{(t)}$ where $i$ is $R_A$'s $i$th antenna, $j$ is $R_B$'s $j$th antenna, and $t$ is the given discrete timestep. Note that as long as we use only our locally collected $z_{ij}^{(t)}$ measurements, we will not require any additional information be exchanged between agents. 

\subsection{Calibrated Range Measurements}
\label{sec:calibrated-measurements}

As noted in Section \ref{sec:measurement-bias-noise}, raw UWB relative range measurements $z_{ij}^{(t)}$ are subject to biases between pairs of antennas $\mu_{ij}$ as well as general non-Gaussian noise. Thus, we can improve quality and robustness of our UWB relative range measurement by: (1) Performing a one-time calibration process to measure the consistent measurement bias $\mu_{ij}$ between nodes $i$ and $j$ respectively (implementation discussed in Section \ref{sec:calibration}). (2) Smoothing sequential UWB relative range measurements with a simple moving average filter,  which should make the signal more closely track the mean while introducing a slight signal delay.

Let $\hat{z}_{ij}^{(t)}$ be a calibrated relative range measurement,
\begin{align}
\hat{z}_{ij}^{(t)} = -\mu_{ij} + \frac{1}{W} \sum_{k=0}^{W-1} z_{ij}^{(t-k)}.
\label{eq:z-hat}
\end{align}
Here we are effectively running a moving average filter over the $W$ most recent $z_{ij}$ range measurements and subtracting out the mean bias $\mu_{ij}$. The choice of $W$ can be selected to trade-off between noise robustness and signal delay, but choosing it to be too large will make rapid relative yaw maneuvers unobservable (i.e., an in-place $360^\circ$ spin within a single $W$ period would appear as if the sensor did not move).

\subsection{2D Formulation - Simple Trilateration on $(x,y,\theta)$}
\label{sec:2d-formulation}
Let $R_A$ and $R_B$ be operating in 2D space, making $R_B$'s relative pose be described as $\mathbf{x} = [x,y,\theta]^T$, where $x,y,\theta$ are the relative $x$-coordinate, $y$-coordinate, and heading (yaw) respectively. Let each robot be equipped with $N \geq 3$ UWB relative range sensors and have their antenna arranged in a non-degenerative layout. Consider the nonlinear least square (NLLS) trilateration pose estimation problem:
\begin{align}
\argmin_{\mathbf{x}} &
\sum_{i \in S} \sum_{j \in S}
\Big(d_{ij}(\mathbf{x}) - \hat{z}_{ij}^{(t)}\Big)^2 \label{eq:2d-opt}
\end{align}
This formulation can be thought of as our baseline implementation that will be augmented in Section \ref{sec:2d-weight}.

\subsection{2D Formulation - Antenna Weighting}
\label{sec:2d-weight}

As noted in Section \ref{sec:antenna}, antennas provide unreliable range measurements when obstructed by another antenna. Since these obstructions are reliably predictable given a specific hardware layout and state, we want to devalue NLLS terms involving an obstructed antenna. This approach is preferable since we are encoding our a priori system knowledge directly into our optimization problem, while more general techniques, such as the use of a Huber loss function,
%or Truncated Least Squares
rely on rejecting data based on general outlier criteria. Our approach can be thought of as an analogue to a Maximum Likelihood Estimator formulation where a measurement covariance matrix is specified as a function of state.

Consider the $2\pi$ periodic weight function $w(\psi)$, specified here on the angular interval $\psi \in [-\pi,\pi]$, and shown plotted in Figure \ref{fig:weight-func}:
\begin{equation}
w(\psi) = \begin{cases}
1 &  \rho \leq |\psi| \leq \pi \\
\frac{1}{2}\cos \left( \frac{\pi (\psi+\rho)}{(\rho-\sigma)}  \right) + \frac{1}{2} & -\rho < \psi < -\sigma \\
\frac{1}{2}\cos \left( \frac{\pi (\psi-\sigma)}{(\rho-\sigma)} + \pi  \right) + \frac{1}{2} & \sigma < \psi < \rho \\
0 & 0 \leq |\psi| \leq \sigma\\
\end{cases}
\label{eq:w}
\end{equation}
where $\sigma$ and $\rho$, related by $0 \leq \sigma \leq \rho \leq \pi$, are predefined constants defining the ``stop-band" end angle and ``pass-band" begin angle respectively. This function can be thought of as a piecewise step function with a smooth transition between high and low values, in the form of re-scaled and shifted $\cos$ segments. Compared to a standard step function, this weight function is differentiable, a useful property for optimization. This function will serve as a primitive for the more specialized weight functions developed in Eqs.~\ref{eq:w_a}, \ref{eq:w_b}, and \ref{eq:2d-opt-w}.

Although this overall formulation is agnostic to exact sensor numbers and layout, we will develop the remainder of this section using the experimental hardware setup described in Section \ref{sec:hardware-robot}. Let our 2D robots have an antenna layout as shown in Figure \ref{fig:robot} and described by the expression:
\begin{equation}
\label{eq:2d-ant}
\boldsymbol{\alpha}_{k}(\mathbf{T}) = \begin{bmatrix}
x+R\cos(\frac{\pi}{2}(k - 1) + \theta) \\
y+R\sin(\frac{\pi}{2}(k - 1) + \theta)
\end{bmatrix}
\end{equation}
where relative pose $T$ has relative components $x,y,\theta$ and $R = 0.35$ m.

\begin{figure}
\centering
\includegraphics[width=.95\linewidth]{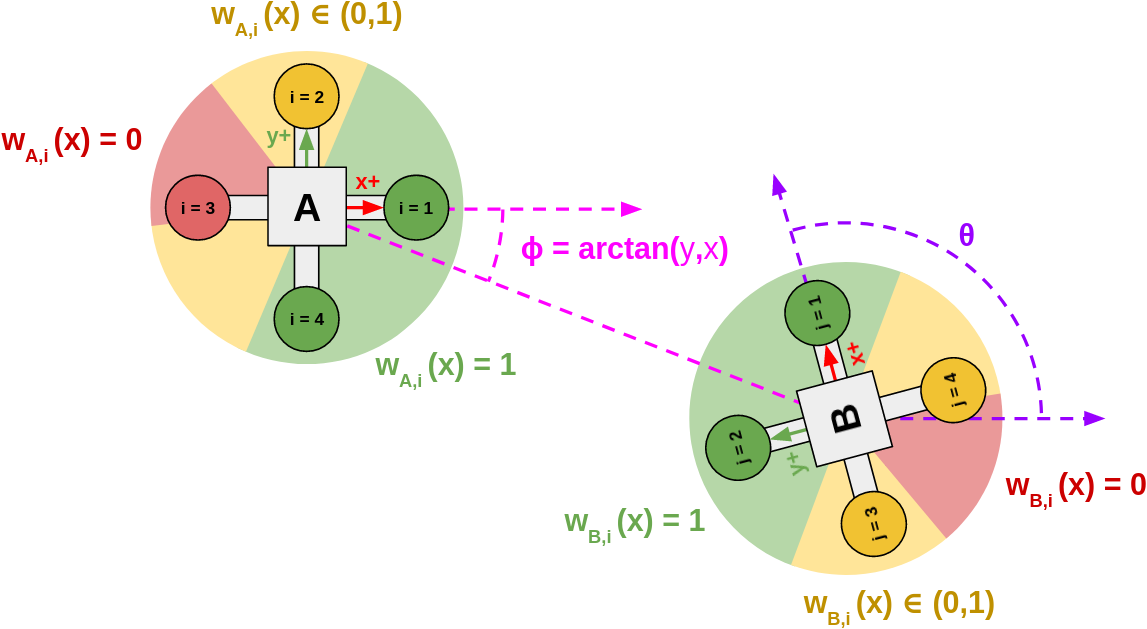}
\caption{Visualization of weighting functions $w_{A,i}(\mathbf{x})$ and $w_{B,i}(\mathbf{x})$ when $\mathbf{x} = [3,-1,100^{\circ}]^T$. Here, the green, yellow, and red colors correspond to the colored weight segments of $w(\psi)$ in Figure \ref{fig:weight-func}. In this scenario, optimization terms involving $R_A$'s $i=3$ antenna are ignored, while terms involving $R_A$'s $i=2$ antenna and/or $R_B$'s $j=3$ or $j=4$ antennas are devalued.}
\label{fig:antenna-ignore}
\end{figure}

Based on our hardware layout, consider the specialized weight functions $w_{A,i}(\mathbf{x})$ and $w_{B,j}(\mathbf{x})$ for discounting nonlinear least squared terms involving $R_A$'s $i$th antenna and $R_B$'s $j$th antenna respectively:
\begin{align}
w_{A,i}(\mathbf{x}) &= w(\arctan(y,x) - \frac{\pi}{2}(i+1)) \label{eq:w_a} \\
w_{B,j}(\mathbf{x}) &= w(\theta - \arctan(y,x) - \frac{\pi}{2}(j-1)) \label{eq:w_b}
\end{align}
Note that $w_{A,i}(\mathbf{x})$ and $w_{B,j}(\mathbf{x})$ are just phase shifted versions of $w(\psi)$ based on the components in $\mathbf{x} = [x,y,\theta]^T$. As shown in Figures \ref{fig:z11-err-vs-heading}-\ref{fig:z14-err-vs-heading}, these weight functions are aligned so that they devalue measurements involving antennas on the ``far" side of either robot. See Figure \ref{fig:antenna-ignore} for a visual example.

%\mytodo{See Figure \ref{} for visual description of these equations. [In figure description, include intuition that $w_{A,i}(\mathbf{x})$ only depends on $x,y$ while $w_{B,j}(\mathbf{x})$ depends on $x,y,\theta$.]}

For convenience, we can then combine $w_{A,i}(\mathbf{x})$ and $w_{B,j}(\mathbf{x})$ into a single weighting
$w_{ij}(\mathbf{x}) = w_{A,i}(\mathbf{x}) \cdot w_{B,j}(\mathbf{x})$.
%and visualize in in Figure \ref{} \mytodo{Heat plot of $w_{ij}(\mathbf{x})$ where $\mathbf{x}$ is parameterized by $(\phi,\theta)$}.
Augmenting Eq.~\ref{eq:2d-opt} with this weighting yields
\begin{equation}
%\label{eq:2d-opt-r}
\argmin_{\mathbf{x}}\sum_{i \in \{1,\ldots,N\}} \sum_{j \in \{1,\ldots,N\}} 
w_{ij}(\mathbf{x}) \Big( d_{ij}(\mathbf{x}) - \hat{z}_{ij}^{(t)} \Big)^2 .
\label{eq:2d-opt-w}
\end{equation}
When using the parameters $\sigma = 30^\circ$ and $\rho = 90^\circ$, this means that, at any given time, at most $7$ of $16$ measurements between a pair of robots can be ignored (i.e. four measurements coming from $R_A$'s ignored antenna, four measurements coming from $R_B$'s ignored antenna, with one measurement being ignored twice). Although with reliable measurements only three antennas per agent are needed to have a fully observable 2D system, the redundant fourth antenna allows us to entirely ignore the obstructed measurements within a given pair of agents while maintaining full 2D observability.

\subsection{Optimization Initialization}
\label{sec:initialization}

When computationally solving the nonlinear problems in Eq.~\ref{eq:2d-opt} or \ref{eq:2d-opt-w}, the optimizer requires initial relative pose $\mathbf{x}_0$. Although we cannot guarantee convexity on either equation, as sufficiently erroneous measurements $z_{ij}^{(t)}$ can make either equation behave irregularly, typically we observe that optimizing Eq.~\ref{eq:2d-opt} yields the same result regardless of selected the $\mathbf{x}_0$, while Eq.~\ref{eq:2d-opt-w} very much depends on the selected $\mathbf{x}_0$. To address this, in practice we perform a two staged optimization, i.e. we initialize Eq.~\ref{eq:2d-opt-w} with $\mathbf{x}_{res}$, which is the result of solving Eq.~\ref{eq:2d-opt} when initialized at $\mathbf{x}_0 = \mathbf{0}$.

To demonstrate the utility of this two-step process, a simulation was written in which ground truth state $\mathbf{x}_{gt}$ was sampled from the uniform distribution $[\mathcal{U}_{[-5,5]},\mathcal{U}_{[-5,5]},\mathcal{U}_{[0,360]}]^T$ such that $\|[x,y]^T\|_2 \geq 1$ and $z_{ij}$ was sampled by adding $\mathcal{N}(0,0.2)$ to the ground truth distances. Table \ref{tab:sim} shows the results after $10000$ trials, which highlight that solving Eq. \ref{eq:2d-opt} effectively finds the same local minimum whether initialized with $\mathbf{0}$ or $\mathbf{x}_{gt}$, but Eq.~\ref{eq:2d-opt-w} often finds the ``wrong" local minimum if initialized with $\mathbf{0}$ and the ``correct" local minimum if initialized with $\mathbf{x}_{res}$ (i.e., the proposed two-step process).

\begin{table}[t]
\caption{Comparison of Mean Difference Predicted Position (MDPP) and Mean Difference Predicted Absolute Heading (MDPAH) over $10000$ trials for different optimization equations initialized with various $\mathbf{x}_0$. Note that MDPP and MDPAH are indicative of how the optimization results differ between the selected initial conditions.}

\resizebox{\linewidth}{!}{%
\begin{tabular}{|c|c||c|c|}
\hline
Opt. Eq. & Compared $\mathbf{x}_0$ & MDPP [m] & MDPAH [deg] \\
\hline \hline
Eq. \ref{eq:2d-opt} & $\mathbf{0}$ vs $\mathbf{x}_{gt}$ & 0.002 & 0.067 \\
%\hline
Eq. \ref{eq:2d-opt-w} & $\mathbf{0}$ vs $\mathbf{x}_{gt}$ & 1.109 & 41.704 \\
%\hline
Eq. \ref{eq:2d-opt-w} & $\mathbf{x}_{res}$ vs $\mathbf{x}_{gt}$ & 0.018 & 0.884 \\
\hline
\end{tabular}%
}
\label{tab:sim}
\end{table}

%Four antenna is the robust version of three antenna.

%While these nonlinear weighting functions improve performance, they also impact the optimization process because experience has shown that much better initial guess are needed for the solution to converge.  This problem is not uncommon for highly nonlinear problems, but in this case it turns out that using a two-stage optimization, with the first stage solved using $w_{ij}=1$ yields a good-enough initial guess to solve the optimization in Eq.~\ref{eq:2d-opt-r}.  Details are provided ...

% Optimization is seeded with simple trilateration approach (i.e. when weight values are constantly 1)
\begin{figure*}
        \captionsetup[subfigure]{aboveskip=-1pt,belowskip=-1pt}

%    \centering
\hspace*{-.25in}
    \begin{multicols}{2}
    \begin{subfigure}{.7\linewidth}
    \includegraphics[width=\linewidth]{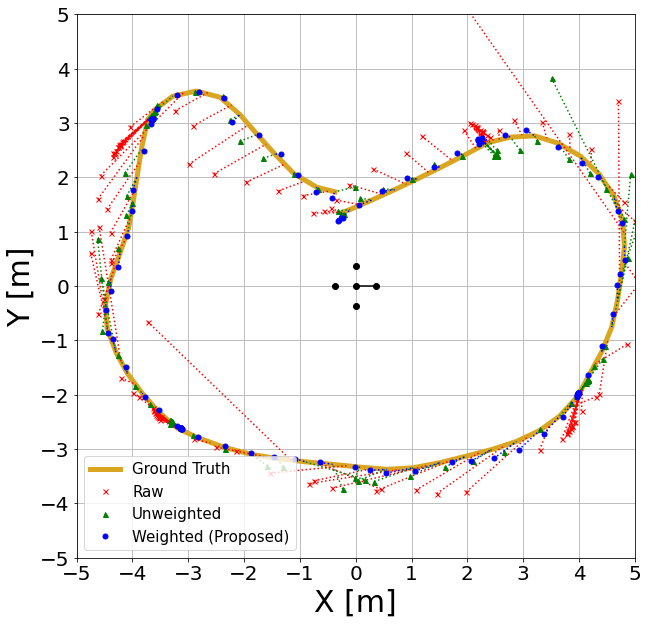}
    \caption{Estimated trajectory of $R_B$ wrt $R_A$}
    \label{fig:bean-traj}
    \end{subfigure}
    \par

\hspace*{-.75in}
    \begin{subfigure}{1.1\columnwidth}
    \includegraphics[width=\linewidth]{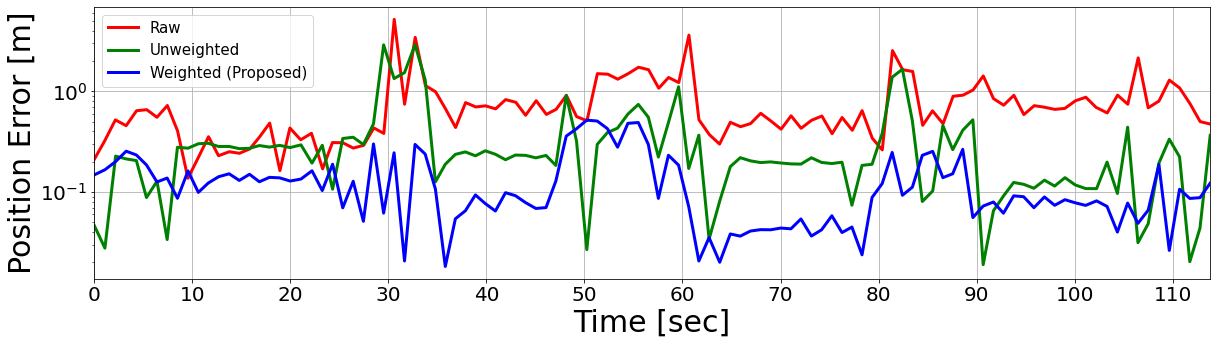}
    \caption{Log-Scale Position Error vs Time}
    \label{fig:bean-pos-err}
    \end{subfigure}
\hspace*{-.425in}
    \begin{subfigure}{1.1\columnwidth}
    \includegraphics[width=\linewidth]{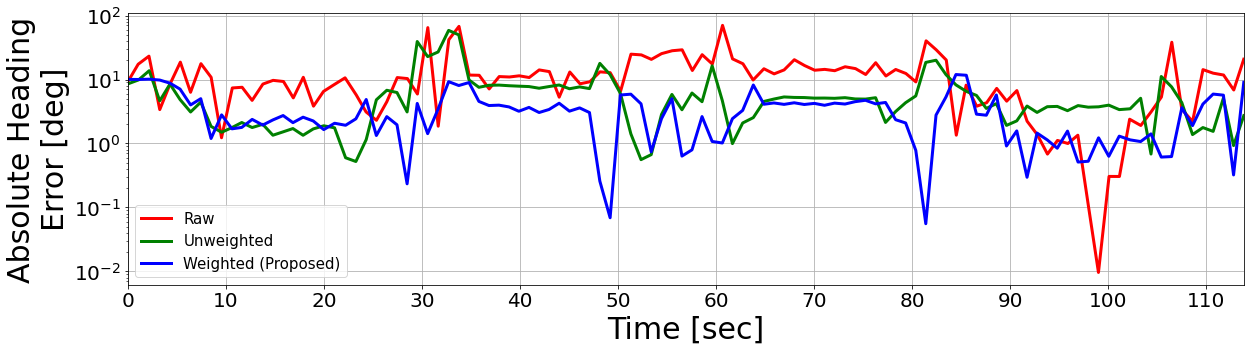}
    \caption{Log-Scale Absolute Heading Error vs Time}
    \label{fig:bean-heading-err}
    \end{subfigure}
    \end{multicols}
\vspace*{-.15in}

    \caption{Results from experimental trial where $R_B$ is manually driven in a kidney bean shape while pausing briefly throughout the trajectory. To prevent visual clutter, only estimations from the \texttt{Raw}, \texttt{Unweighted}, and \texttt{Weighted (Proposed)} algorithms are shown alongside the mocap ground truth. Where applicable the parameters $\sigma = 30^{\circ}$, $\rho = 90^{\circ}$, and $W = 50$ (i.e. $1$ sec of samples since measurements occur at 50Hz) are used.
    (a) Shows each algorithm's predicted position estimate (at 1Hz) and draws an error line between each estimated point and its corresponding ground truth point. (b) Shows each algorithm's log-scale positional error verses time for the given trial. (c) Similarly shows each algorithm's log-scale absolute heading error for the given trial.
    Examining the plots (b) and (c), we see \texttt{Weighted (Proposed)} provides the best estimate of position and heading a vast majority of the times within the trajectory. See specific calculated mean, max, and standard deviation values, as well as all other results, in Tables \ref{tab:pos-err} and \ref{tab:heading-err}.}
    \label{fig:trial-kidney-bean}
\end{figure*}

\section{Experimental Results}
\label{sec:results}

\subsection{Experimental Setup \& Implementation}
\label{sec:implementation}
Experimental trials were conducted using the hardware in Section \ref{sec:hardware-robot} and the algorithms in Section \ref{sec:problem}. All code was written using Python 3 and ROS \cite{quigley2009ros}. Optimization was done using \texttt{scipy.optimize.minimize} with the \texttt{trust-constr} method \cite{2020SciPy-NMeth}.
Calibrated measurements $\hat{z}_{ij}^{(t)}$ are generated at 50Hz, the sensor operating rate, and sampled by the optimization code as needed. The optimization implementation runs in real-time at approximately 10Hz on an Intel i7-6700 with 8GB of RAM. This could easily be sped up by switching to a C++ implementation.
%migrating away from Python and reducing intermediary logging steps.
Collected UWB data is post-processed with several alternative algorithms to compare results. When applicable, we use parameters $\sigma = 30^{\circ}$, $\rho = 90^{\circ}$, and $W = 50$ (measurements sampled at 50Hz). Specifically, these algorithms are:
\begin{itemize}
\item \texttt{Raw}: Optimizes Eq. \ref{eq:2d-opt}, but uses the ``raw" $z_{ij}^{(t)}$ measurement instead of the calibrated $\hat{z}_{ij}^{(t)}$.
\item \texttt{Shift only}: Refers to Eq. \ref{eq:2d-opt}, but uses the $z_{ij}^{(t)} - \mu_{ij}$ instead of the calibrated $\hat{z}_{ij}^{(t)}$.
\item \texttt{MovingAvg only}: Refers to Eq. \ref{eq:2d-opt}, but uses $\frac{1}{W} \sum_{k=0}^{W-1} z_{ij}^{(t-k)}$ instead of the calibrated $\hat{z}_{ij}^{(t)}$.
\item \texttt{Unweighted}: Refers to Eq. \ref{eq:2d-opt} as written.
\item \texttt{Weighted (Proposed)}: Optimizes Eq.~\ref{eq:2d-opt-w}.
\end{itemize}
Comparing the results of \texttt{Raw}, \texttt{Shift only}, and \texttt{MovingAvg only} with \texttt{Unweighted} will clearly show the benefits of using the calibrated $\hat{z}_{ij}^{(t)}$. Similarly, comparing the results of \texttt{Unweighted} and \texttt{Weighted} shows the benefits of adding the NLLS weighting $w_{ij}(\mathbf{x})$.

\subsection{Calibration}
\label{sec:calibration}
Calibration means $\mu_{ij}$ were found by placing $R_A$ and $R_B$ a known distance apart ($3$ m) and rotating $R_B$ in place at approximately $60$ deg/sec. After $R_B$ completed a full revolution, $R_A$ was rotated approximately $30^\circ$ before $R_B$ resumed rotating. The process took approximately $90$ seconds and, afterwards, $\mu_{ij}$ values were calculated by averaging $\tilde{z}_{ij}$ while omitting regions of antenna obstructions spikes, similar to Figures \ref{fig:z11-err-vs-heading}-\ref{fig:z14-err-vs-heading}. The following $\mu_{ij}$ values were computed:
%\small
\begin{align*}
\mu_{11} &= 0.268 & \mu_{12} &= 0.266 & \mu_{13} &= 0.277 & \mu_{14} &= 0.230 \\
\mu_{21} &= 0.093 & \mu_{22} &= 0.112 & \mu_{23} &= 0.227 & \mu_{24} &= 0.188 \\
\mu_{31} &= 0.046 & \mu_{32} &= 0.018 & \mu_{33} &= 0.170 & \mu_{34} &= 0.078 \\
\mu_{41} &= 0.041 & \mu_{42} &= 0.065 & \mu_{43} &= 0.178 & \mu_{44} &= 0.095
\end{align*}
%\normalsize
Although these values are specific to our hardware, they show the significance of these biases given the current operating scale as well as how much variation there is between pairs of sensors (i.e.,~as much as $25.9$ cm between our observed best and worst pair).

\subsection{Trials}
\label{sec:trials}

Several experiments were run with different trajectories and durations. In all trials the Turtlebots were driven at the maximum velocity and rotation rate ($1$ m/s of $1$ rad/s respectively). The specific trials were:
\begin{itemize}
\item \texttt{rot-cw/rot-ccw}: $R_A$ is kept stationary. $R_B$ is placed a fixed distance away and rotated clockwise/counter-clockwise in place.
\item \texttt{traj-cw/traj-ccw}: $R_A$ is kept stationary. $R_B$ is manually driven in a circular trajectory about $R_A$ in a clockwise/counter-clockwise direction.
\item \texttt{kidney-bean}: $R_A$ is kept stationary. $R_B$ is manually driven in a kidney bean shape while pausing briefly throughout the trajectory (See Figure \ref{fig:trial-kidney-bean}).
\item \texttt{box}: $R_A$ is kept stationary. $R_B$ is manually driven in an approximately $8$ m by $6$ m rectangle.
\item \texttt{both-move}: Both $R_A$ and $R_B$ are manually driven arbitrarily within within a $10$ m by $10$ m space without getting within $1$ m of each other.
\end{itemize}
All results are compiled into Tables \ref{tab:pos-err} and \ref{tab:heading-err}, showing the mean, max, and standard deviation of the position and absolute heading errors respectively.

\begin{table*}[h!]
\caption{Evaluation of positional error in meters between algorithms and datasets.}

\resizebox{\textwidth}{!}{%
\begin{tabular}{|l||c|c|c|c|c|c|c|c|c|c|c|c|c|c|c|c|c|c|c|c|c|c|}
\hline
\multirow{2}{3.0cm}{\textbf{Position Error [m]}} & \multicolumn{21}{|c|}{Scenario} \\
\cline{2-22} & \multicolumn{3}{|c|}{rot-cw} & \multicolumn{3}{|c|}{rot-ccw} & \multicolumn{3}{|c|}{traj-cw} & \multicolumn{3}{|c|}{traj-ccw} & \multicolumn{3}{|c|}{kidney-bean} & \multicolumn{3}{|c|}{box} & \multicolumn{3}{|c|}{both-move} \\
\hline \textbf{Method} & \textbf{Mean} & \textbf{Max} & \textbf{Std} & \textbf{Mean} & \textbf{Max} & \textbf{Std} & \textbf{Mean} & \textbf{Max} & \textbf{Std} & \textbf{Mean} & \textbf{Max} & \textbf{Std} & \textbf{Mean} & \textbf{Max} & \textbf{Std} & \textbf{Mean} & \textbf{Max} & \textbf{Std} & \textbf{Mean} & \textbf{Max} & \textbf{Std} \\
\hline \hline
Raw \rule{0pt}{7pt} & 0.44 & 0.98 & 0.16 & 0.38 & 0.65 & 0.08 & 0.98 & 5.71 & 0.88 & 0.59 & 1.54 & 0.30 & 0.76 & 5.20 & 0.68 & 0.93 & 4.99 & 0.70 & 0.82 & 3.50 & 0.54 \\
%\hline
Shift only & 0.27 & 0.72 & 0.15 & 0.19 & 0.46 & 0.09 & 0.70 & 5.27 & 0.89 & 0.27 & 1.27 & 0.25 & 0.47 & 4.77 & 0.73 & 0.44 & 4.61 & 0.69 & 0.42 & 3.53 & 0.48 \\
%\hline
MovingAvg only & 0.35 & 0.53 & 0.05 & 0.34 & 0.47 & 0.05 & 0.87 & 5.08 & 0.65 & 0.55 & 1.41 & 0.27 & 0.68 & 3.40 & 0.47 & 0.91 & 3.57 & 0.59 & 0.78 & 2.64 & 0.42 \\
%\hline
Unweighted (Eq. \ref{eq:2d-opt}) & 0.14 & 0.25 & \textbf{0.02} & 0.13 & \textbf{0.19} & \textbf{0.02} & 0.62 & 4.37 & 0.70 & 0.22 & 1.02 & 0.19 & 0.36 & 2.91 & 0.44 & 0.45 & 3.29 & 0.54 & 0.37 & 2.87 & 0.42 \\
%\hline
Weighted (Eq. \ref{eq:2d-opt-w}) \textbf{[Proposed]} & \textbf{0.09} & \textbf{0.21} & 0.06 & \textbf{0.09} & 0.21 & 0.05 & \textbf{0.20} & \textbf{1.41} & \textbf{0.18} & \textbf{0.21} & \textbf{0.86} & \textbf{0.16} & \textbf{0.13} & \textbf{0.52} & \textbf{0.11} & \textbf{0.21} & \textbf{0.96} & \textbf{0.17} & \textbf{0.29} & \textbf{2.70} & \textbf{0.32} \\
\hline
\end{tabular}%
}
\label{tab:pos-err}
\end{table*}

\begin{table*}[h!]
\caption{Evaluation of absolute heading error in degrees between algorithms and datasets.}

\resizebox{\textwidth}{!}{%
\begin{tabular}{|l||c|c|c|c|c|c|c|c|c|c|c|c|c|c|c|c|c|c|c|c|c|c|}
\hline
\multirow{2}{3.5cm}{\textbf{Abs Heading Error [deg]}} & \multicolumn{21}{|c|}{Scenario} \\
\cline{2-22} & \multicolumn{3}{|c|}{rot-cw} & \multicolumn{3}{|c|}{rot-ccw} & \multicolumn{3}{|c|}{traj-cw} & \multicolumn{3}{|c|}{traj-ccw} & \multicolumn{3}{|c|}{kidney-bean} & \multicolumn{3}{|c|}{box} & \multicolumn{3}{|c|}{both-move} \\
\hline \textbf{Method} & \textbf{Mean} & \textbf{Max} & \textbf{Std} & \textbf{Mean} & \textbf{Max} & \textbf{Std} & \textbf{Mean} & \textbf{Max} & \textbf{Std} & \textbf{Mean} & \textbf{Max} & \textbf{Std} & \textbf{Mean} & \textbf{Max} & \textbf{Std} & \textbf{Mean} & \textbf{Max} & \textbf{Std} & \textbf{Mean} & \textbf{Max} & \textbf{Std} \\
\hline \hline
Raw \rule{0pt}{7pt}& 9.82 & 21.81 & 5.53 & 10.10 & 22.43 & 6.05 & 16.82 & 90.29 & 15.09 & 12.44 & 35.78 & 9.38 & 12.06 & 70.33 & 12.17 & 12.11 & 65.86 & 10.15 & 11.35 & 72.69 & 9.33 \\
%\hline
Shift only & 6.38 & 16.44 & 4.19 & 6.45 & 14.20 & 4.09 & 12.40 & 90.41 & 15.52 & 6.99 & 31.55 & 6.44 & 8.10 & 80.20 & 12.72 & 6.34 & 64.33 & 9.76 & 6.97 & 78.23 & 9.46 \\
%\hline
MovingAvg only & 7.05 & 14.70 & 4.63 & 7.37 & 17.59 & 4.68 & 14.48 & 77.47 & 11.49 & 11.27 & 35.41 & 8.70 & 10.32 & 57.96 & 8.73 & 11.55 & 50.65 & 9.01 & 11.11 & 55.54 & \textbf{7.15}\\
%\hline
Unweighted (Eq. \ref{eq:2d-opt}) & \textbf{3.31} & \textbf{8.28} & \textbf{1.99} & \textbf{2.88} & \textbf{9.78} & \textbf{2.13} & 10.77 & \textbf{72.66} & 11.65 & \textbf{5.44} & \textbf{25.92} & \textbf{4.79} & 6.15 & 58.70 & 8.38 & 5.99 & 51.71 & 7.91 & 6.38 & \textbf{53.30} & 7.64\\
%\hline
Weighted (Eq. \ref{eq:2d-opt-w}) \textbf{[Proposed]} & 4.12 & 10.78 & 3.11 & 4.43 & 10.53 & 3.34 & \textbf{5.88} & 75.56 & \textbf{8.09} & 5.84 & 31.01 & 6.05 & \textbf{3.95} & \textbf{11.97} & \textbf{2.98} & \textbf{4.22} & \textbf{13.31} & \textbf{3.03} & \textbf{5.45} & 104.00 & 9.97 \\
\hline
\end{tabular}%
}
\label{tab:heading-err}
\end{table*}

\subsection{Interpreting Results}
\label{sec:analysis}
The results in Tables \ref{tab:pos-err} and \ref{tab:heading-err} show that \texttt{Weighted} and \texttt{Unweighted} consistently outperform the other approaches; this makes clear the advantage of the calibrated $\hat{z}_{ij}^{(t)}$ over the raw $z_{ij}^{(t)}$ or the individual mean shift/moving average corrections. Additionally, \texttt{Weighted} consistently outperforms \texttt{Unweighted} in mean positional error. Next, while \texttt{Unweighted} outperforms \texttt{Weighted} in a few select metrics in the simpler \texttt{rot-*} and \texttt{traj-*} trials, this is only by relatively small margins (i.e.,~at most $4$ cm or $5.1$ deg respectively). When considering the more challenging \texttt{kidney-bean} and \texttt{box} trajectories, we see \texttt{Weighted} substantially outperforms \texttt{Unweighted} in all metrics. Finally, when examining \texttt{both-move}, we see \texttt{Weighted} outperforms all other methods in the three positional error metrics as well as mean absolute heading error. The spike observed in \texttt{Weighted}'s other two heading metrics appears to be the result of rapid relative yawing, possible when $R_A$ and $R_B$ yaw simultaneously at max speed in opposite directions.

%substantially in the more challenging \texttt{kidney-bean} and \texttt{box} trajectories (e.g. by $23$ cm and $24$ cm respectively).
%Thus, while not a strict improvement over \texttt{Unweighted} in the simpler trials, we see \texttt{Weighted} has consistent and %substantial improvements in the most challenging trials, as well as consistent improvements in mean positional error (debatably the %most important metric) across the board.

\subsection{Comparison to Literature}
\label{sec:comparison}
Our \texttt{box} trial was designed so that it can be compared to a similar experiment in \cite{DBLP:conf/iros/Cao0YANMT21}. Note that both experiments used two Turtlebots (one stationary, one moving), each equipped with four Nooploop LinkTrack P UWB modules separated by an approximately $70$ cm baseline, and performed relative pose estimation while traversing an approximately $8$ m by $6$ m rectangle. When comparing our results to the reported results in \cite{DBLP:conf/iros/Cao0YANMT21}, our proposed approach, \texttt{Weighted}, achieved slightly better position error ($0.21$ m vs $0.25$ m), but with slightly worse mean heading error ($4.22$ deg vs $2.02$ deg) and standard deviations (our $0.17$ m and $3.03$ deg vs their $0.09$ m $1.47$ deg respectively). 
Thus, our approach is competitive with the performance in \cite{DBLP:conf/iros/Cao0YANMT21}, 
but that work assumes access to \textbf{continuously transmitted odometry estimates, whereas our approach does not}.
%%while not relying on the constant transmission of data, and thus getting similar results with significantly less information.
\section{Conclusion}
\label{sec:conclusion}

We proposed a multi-agent 2D relative pose localization approach that does not rely on any external infrastructure or data exchange between agents, just multiple locally collected UWB range measurements. By integrating a priori knowledge about our observed measurement biases and obstruction patterns, we achieve competitive results with works such as \cite{DBLP:conf/iros/Cao0YANMT21}, but without needing continuously transmitted odometry between agents.

In the future, we plan to extend this approach to 3D pose, while maintaining minimalist communication requirements. We will also investigate increasing the team to have more agents, better modeling of the range measurement noise, estimating our pose covariance through more powerful optimization libraries, and taking a more rigorous approach to our observability analysis. Finally, we will directly integrate this system into full-scale multi-agent collaborative SLAM pipeline, like Kimera-Multi \cite{DBLP:conf/icra/ChangTHC21}.

\iffalse
We proposed a multi-agent relative pose localization approach that does not rely on any external infrastructure or data exchange between agents, just multiple locally collected UWB range measurements. By not relying on inter-agent data exchange, our algorithm has a niche use as a informative subsystem in a larger multi-agent collaborative SLAM pipeline.

By integrating a priori knowledge about our agents' hardware layouts, we improve the performance of our NLLS trilateration approach with a term weighting function that varies with respect to the estimated relative pose. Although our implementation targets our specific hardware layout, the approach put forth is general and adaptable to other platforms with alternative layouts and obstructions.

Our results provide slightly improved mean positional accuracy and remains competitive with respect to other metrics in the similar work \cite{DBLP:conf/iros/Cao0YANMT21}, but without necessitating the continuous transmission of agent odometry between agents.

We plan to extend this approach to 3D (6-DoF pose), while maintaining minimalist communication requirements. We will also investigate increasing the team to have more agents, better modeling of the range measurement noise, estimating our pose covariance through more powerful optimization libraries, and a rigorous observability analysis. Finally, we will directly integrate this system into a full-scale multi-agent SLAM pipeline, such as Kimera-Multi \cite{DBLP:conf/icra/ChangTHC21}.
\fi

\balance

% \nocite{*}
% Future: Look into proper way to truncate Nebula and Scipy authors
% Currently doing first 5 authors + et al.
%\bibliographystyle{abbrv}
\bibliographystyle{unsrt}
\bibliography{mybib}{}
%\bibliographystyle{plain}
%\printbibliography

\appendices
%\input{src/09_Appendix}

%\clearpage
%\input{src/08_Temp}

\end{document}